\documentclass[nomathfonts]{aipproc} 
\layoutstyle{6s}
\usepackage{amsfonts}
\usepackage{amsmath}
\usepackage{amssymb}
\usepackage{graphicx}

\newcommand{\prob}{\mathcal{P}}
\newcommand{\dird}{\mathcal{D}}

\title{Online Learning in Discrete Hidden Markov Models}
\author{Roberto Alamino}
 {
  address = {Neural Computing Research Group,\linebreak 
  Aston University \linebreak  
  Aston Triangle, Birmingham, B4 7ET, United Kingdom},
  email = {alaminrc@aston.ac.uk}
 }
\author{Nestor Caticha}
 {
  address = {Instituto de F\'isica,\linebreak 
             Universidade de S\~ao Paulo,\linebreak
             CP 66318 S\~ao Paulo, SP, CEP 05389-970 Brazil},
  email = {nestor@if.usp.br}
}
\date{}

\begin{abstract}
We present and analyse three online algorithms for learning in
discrete Hidden Markov Models (HMMs) and compare them with the 
Baldi-Chauvin Algorithm. Using the Kullback-Leibler divergence as a measure of 
generalisation error we draw learning 
curves in simplified situations. The performance 
for learning drifting concepts of one of the presented algorithms is analysed and
compared with the Baldi-Chauvin algorithm in the same situations.  
A brief discussion about learning and symmetry breaking based on our 
results is also presented.
\\ ~\\
{\bf Key Words: }
HMMs, Online Algorithm, Generalisation Error, Bayesian Algorithm.
\end{abstract}

\begin{document}
\maketitle

\section{Introduction}

\textit{Hidden Markov Models} (HMMs) \cite{Ephraim02,Rabiner89} are extensively studied machine learning 
models for time series with several applications in fields 
like speech recognition \cite{Rabiner89}, bioinformatics \cite{Baldi01,Durbin98} and LDPC codes 
\cite{Frias04}. They consist of a Markov chain of non-observable \emph{hidden states} 
$q_t\in S$, $t=1,...,T$, $S=\{s_1, s_2,...,s_n\}$, with initial probability vector $\pi_i=\prob(q_1=s_i)$ and 
transition matrix $A_{ij}(t)=\prob(q_{t+1}=s_j|q_t=s_i)$, $i,j=1,..,n$. At discrete times $t$, each $q_t$ emits an
\emph{observed state} $y_t\in O$, $O=\{o_1,..., o_m\}$, with emission probability matrix 
$B_{i\alpha}(t)=\prob(y_t=o_\alpha|q_t=s_i)$, $i=1,...,n$, $\alpha=1,...,m$, which are the 
actual observations of the time series represented, from time $t=1$ to $t=T$, by the \textit{observed sequence} 
$y_1^T=\{y_1, y_2, ..., y_T\}$. The $q_t$'s form the so called \textit{hidden sequence} $q_1^T=\{q_1,q_2,...,q_T\}$.
The probability of observing a sequence $y_1^T$ given $\omega\equiv(\pi,A,B)$ is
\begin{equation}
  \prob(y_1^T|\omega) = \sum_{q_1^T} \prob(y_1)\prob(y_1|q_1)\prod_{t=2}^T \prob(q_{t+1}|q_t)\prob(y_t|q_t).
\end{equation}
 
In the \textit{learning process}, the HMM is fed with a series and adapts its parameters to produce similar ones. 
Data feeding 
can range from \textit{offline} (all data is 
fed and parameters calculated all at once) to \textit{online} (data is fed by parts and partial calculations are made). 

We study a scenario with data generated by a HMM of unknown parameters, an 
extension of the student-teacher scenario from neural networks. The performance, as a function of the number of 
observations, is given 
by how \textit{far}, measured by a suitable criterion, is the student from the teacher. Here we 
use the naturally arising \textit{Kullback-Leibler (KL) divergence} that, although not accessible in 
practice since it needs knowledge of the teacher, is an extension of the idea of 
generalisation error being very informative. 

We propose three algorithms and compare them with the \textit{Baldi-Chauvin Algorithm} (BC) 
\cite{Baldi94}: the \textit{Baum-Welch Online Algorithm} (BWO), an adaptation of the offline 
\textit{Baum-Welch Reestimation Formulas} (BW) \cite{Ephraim02} and, starting from a Bayesian formulation, 
an approximation named \textit{Bayesian Online Algorithm} (BOnA), that can be simplified again without 
noticeable lost of performance to a \textit{Mean Posterior Algorithm} (MPA). BOnA and MPA, 
inspired by  
Amari \cite{Amari} and Opper \cite{Opper98}, are essentially mean field methods \cite{Saad} in which a manifold of 
prior tractable distributions is introduced and the
new datum leads, through Bayes theorem, to a non-tractable posterior. The key step is to take as the
new prior, not the posterior, but the closest distribution (in some sense) in the manifold. 

The paper is organised as follows: first, BWO is introduced and analysed. Next, we 
derive BOnA for HMMs and, from it, MPA. We compare
MPA and BC for drifting concepts. Then, we discuss learning and symmetry 
breaking and end with our conclusions.

\section{Baum-Welch Online Algorithm}

The \textit{Baum-Welch Online Algorithm} (BWO) is an online adaptation of BW where in 
each iteration of BW, $y$
becomes $y^p$, the $p$-th observed sequence. Multiplying the BW increment by a 
learning rate $\eta_{BW}$ we get the update equations for $\omega$
\begin{equation}
  \hat{\omega}^{p+1} = \hat{\omega}^p + \eta_{BW} \hat{\Delta} \omega^p,
\end{equation}
with $\hat{\Delta}\omega^p$ the BW variations for $y^p$. 
The complexity of BWO is polynomial in $n$ and $T$.

\begin{figure}
\centering
\includegraphics[width=5.8cm]{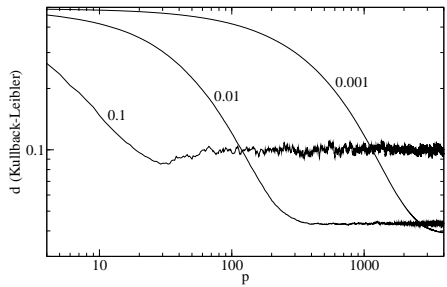}
\caption{Log-log curves of BWO for three different $\eta_{BW}$ indicated next to the curves.}
\label{fBWO01}
\end{figure}

In figure \ref{fBWO01}, the HMM learns sequences generated by a teacher with $n=2$, $m=3$ and $T=2$ 
for different $\eta_{BW}$. Initial students have matrices with all entries set to the 
same value, what we call a \textit{symmetric initial student}. We took averages over 500 random teachers and distances 
are given by the KL-divergence between two HMMs $\omega_1$ and $\omega_2$
\begin{equation}
  \label{DKL_HMM}
  d_{KL}(\omega_1,\omega_2) \equiv \sum_{y_1^T}\prob(y_1^T|\omega
    _1) \ln \left[\frac{\prob(y_1^T|\omega _1)}{\prob(y_1^T|
    \omega _2)}\right]. 
\end{equation}

We see that after a certain number of sequences the HMM stops 
learning, which is particular to the symmetric initial 
student and disappears for a non-symmetric one.

Denoting the variation of the parameters in BC by $\Delta$, in BW by $\hat{\Delta}$, in BWO by $\tilde{\Delta}$, 
and with $\gamma_t(i)\equiv\prob(q_t=s_i|y^p,\omega^p)$, we have to first order in 
$\lambda$ 
\begin{eqnarray}
  \label{BWxBC}
  \Delta \pi_i       &=& \frac{\lambda\eta_{BC}}n\hat{\Delta}\pi_i=
                         \frac{\lambda}n \frac{\eta_{BC}}{\eta_{BW}}
                         \tilde{\Delta}\pi_i,\\
  \Delta A_{ij}      &=& \frac{\lambda\eta_{BC}}n\left[\sum_{t=1}^{T-1}
                         \gamma_t(i)\right]\hat{\Delta}A_{ij}=
                         \frac{\lambda}n \frac{\eta_{BC}}{\eta_{BW}}
                         \left[\sum_{t=1}^{T-1}
                         \gamma_t(i)\right]\tilde{\Delta}A_{ij}, \nonumber\\
  \Delta B_{i\alpha} &=& \frac{\lambda \eta_{BC}}n\left[\sum_{t=1}^T
                          \gamma_t(i)\right]\hat{\Delta}B_{i\alpha}=
			  \frac{\lambda}n\frac{\eta_{BC}}{\eta_{BW}}
                          \left[\sum_{t=1}^T
                          \gamma_t(i)\right]\tilde{\Delta}B_{i\alpha}.
                          \nonumber
\end{eqnarray}

For $\eta_{BW} \approx \lambda \eta_{BC}/n$ and 
small $\lambda$, variations in BC are proportional to those in BWO, but with different 
effective learning rates for each matrix depending on $y^p$. Simulations show that actual values are of the same order 
of approximated ones. 

\section{The Bayesian Online Algorithm}

The Bayesian Online Algorithm (BOnA) \cite{Opper98} uses Bayesian inference to adjust $\omega$ in the HMM using a 
data set $D_P=\{y^1,...,y^P\}$. For each data, the prior distribution is updated by Bayes' theorem. This update
takes a prior from a parametric family and transforms it in a posterior which in general has no longer the same 
parametric form. The strategy used by BOnA is then to project the posterior back into the initial parametric family. In
order to achieve this, we minimise the KL-divergence between the posterior and a distribution in the parametric family. 
This minimisation will enable us to find the parameters of the closest parametric distribution by which we will 
approximate our posterior. The student HMM $\omega$ parameters in each step of the learning process are estimated as the 
means of the each projected distribution.

For a parametric family that has the form $P(x)\propto e^{-\sum_i\lambda_i f_i(x)}$,
which can be obtained by the MaxEnt principle where we constrain the averages over $P(x)$ 
of arbitrary functions $f_i(x)$, minimising the KL-divergence turns out to be equivalent 
to equating the averages $<f_i(x)>$ over $P(x)$ to the average of these functions over the unprojected posterior (our
posterior distribution just after the Bayesian update for the next data).

For HMMs, the vector $\pi$ and each $i$-th row $A^i$ of $A$ and $B^i$ of $B$ are different discrete distributions which we assume independent 
in order to write the factorized distribution
\begin{equation}
  \label{fPrDis2}
  \prob(\omega|u)\equiv \prob(\pi|\rho)\prod_{i=1}^n 
                        \prob(A^i|a^i) \prob(B^i|b^i),
\end{equation}
where $u=(\rho,a,b)$ represents the parameters of the 
distributions.

As each factor is a distribution over probabilities, the natural choice are 
the Dirichlet distributions, which for a $N$-dimensional variable $x$ is 
\begin{equation}
 \dird(x|u)=\frac{\Gamma(u_0)}
              {\prod_{i=1}^N\Gamma(u_i)} \prod_{i=1}^N x_i^{u_i-1}, 
\end{equation}
where $u_0=\sum_i u_i$ and $\Gamma$ is the analytical continuation of the factorial to real numbers. 
These can be obtained from MaxEnt with $f_i(x)=\ln x_i$ \cite{Vlad}:
\begin{equation}
  \int d\mu\, \dird(x) \ln x_i=\alpha_i, \qquad
  d\mu \equiv \delta \left( \sum_i x_i -1 \right) \prod_i \theta(x_i) dx_i.
\end{equation} 

The function to be extremized is
\begin{equation}
  \mathcal{L} = \int d\mu\,\dird\ln\dird + \lambda \left(\int d\mu\,\dird 
                -1\right) + \sum_i\lambda_i\left(\int d\mu\,\dird\ln x_i -
                \alpha_i\right),
\end{equation} 
and with $\delta\mathcal{L}/\delta\dird=0$ we get
the Dirichlet with normalisation $e^{\lambda+1}$ and $u_i=1-\lambda_i$.

Each factor distribution is separately projected by
equating the average of the logarithms in the original posterior $Q$ and in the projected distributions
\begin{eqnarray}
  \label{fDSM}
  \psi(\rho_i) -\psi\left( \sum_j \rho_j\right) &=& \left<\ln\pi_i\right>_Q 
    \equiv \mu_i(\rho),\\
  \psi(a_{ij}) -\psi\left( \sum_k a_{ik}\right) &=& \left<\ln A_{ij}\right>_Q 
    \equiv \mu_{ij}(a),\nonumber\\
  \psi(b_{i\alpha})-\psi\left(\sum_\beta b_{i\beta}\right) &=& 
     \left<\ln B_{i\alpha}\right>_Q \equiv \mu_{i\alpha}(b),\nonumber
\end{eqnarray}
where
$\psi(x) =d\ln\Gamma(x)/dx$ is the digamma function.
We call a set of $N$ equations
\begin{equation}
  \label{digamma}
  \psi(x_i) -\psi\left( \sum_j x_j\right) = \mu_i,\\
\end{equation}
with $i=1,...N$ a \textit{digamma system} in the variables $x_i$ with 
coefficients $\mu_i$. 

Let us call $P^p(\omega)$ the projected distribution after observation of $y^p$, and 
$Q^{p+1}(\omega)$ the posterior distribution (not projected yet) after $y^{p+1}$.
By Bayes' theorem, 
\begin{equation}
  \label{fPD}
  Q^{p+1}(\omega) \propto P^p(\omega) \sum_{q^{p+1}}
                   \prob(y^{p+1},q^{p+1}|\omega). 
\end{equation}

The calculation of $\mu$'s in (\ref{fDSM}) leads to averages over Dirichlets of the 
form \cite{Alamino}
\begin{equation}
  \label{fDDML}
  \mu_i=\left< \left[ \prod_j x_j^{r_j}\right]\ln x_i\right>=
  \frac{
  \Gamma (u_0)}{\prod_j\Gamma (u_j)}\frac{\prod_j\Gamma (u_j+r_j)}{\Gamma
  (u_0+r_0)}[\psi(u_i+r_i) -\psi(u_0+r_0)].
\end{equation}

To solve (\ref{digamma}),
we solve for $x_i$, sum over $i$ with $x_0\equiv\sum_i x_i$ and find numerically, by iterating from an 
arbitrary initial point, the fixed points of the one-dimensional map
\begin{equation}
  x_0^{n+1}=\sum_i\psi^{-1}[\mu _i+\psi(x_0^n)],
\end{equation}
where we found a unique solution except for $\mu_i \approx 0$, which is rare in most applications.

BOnA has a common problem of Bayesian algorithms: the sum over hidden variables makes the complexity scales 
exponentially in $T$. Also, the calculation of several digamma functions is very time consuming. In the following, 
we develop an approximation that runs faster, although still with exponential complexity in $T$.
This is not a problem for we can make $T$ constant and the algorithm will scale polynomially in $n$. 

\section{Mean Posterior Approximation}
The Mean Posterior Approximation (MPA) is a simplification of BOnA inspired in its results for Gaussians, 
where we match first and second moments of posterior and projected distributions. Noting it, instead of 
minimising $d_{KL}$ we match the mean and one of the variances of posterior and projected 
distributions as an approximation, which gives, with hatted variables for reestimated values \cite{Alamino}
\begin{eqnarray}
  \hat{\rho}_i &=& \left<\pi_i\right>_Q
                   \frac{\left<\pi_1\right>_Q-\left<\pi_1^2
                   \right>_Q}{\left<\pi_1^2\right>_Q
                   -\left<\pi_1\right>_Q^2}, \\
  \hat{a}_{ij} &=& \left<a_{ij}\right>_Q
                   \frac{\left<a_{i1}\right>_Q-\left<a_{i1}^2
                   \right>_Q}{\left<a_{i1}^2\right>_Q
                   -\left<a_{i1}\right>_Q^2}, \nonumber\\ 
  \hat{b}_{i\alpha} &=& \left<b_{i\alpha}\right>_Q
                   \frac{\left<b_{i1}\right>_Q-\left<b_{i1}^2
                   \right>_Q}{\left<b_{i1}^2\right>_Q
                   -\left<b_{i1}\right>_Q^2},\nonumber
\end{eqnarray}
with complexity again of order $n^T$, but with heavily reduced real computational
time making it better for practical applications.

\begin{figure}
\centering
\includegraphics[width=5.8cm]{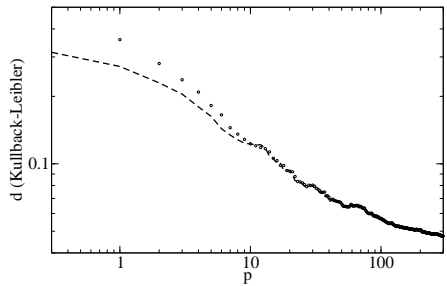}
\caption{Comparison in log-log scale of MPA (dashed line) and 
         BOnA (circles).}
\label{fBWO06}
\end{figure}

\begin{figure}
\centering
\includegraphics[width=13.8cm]{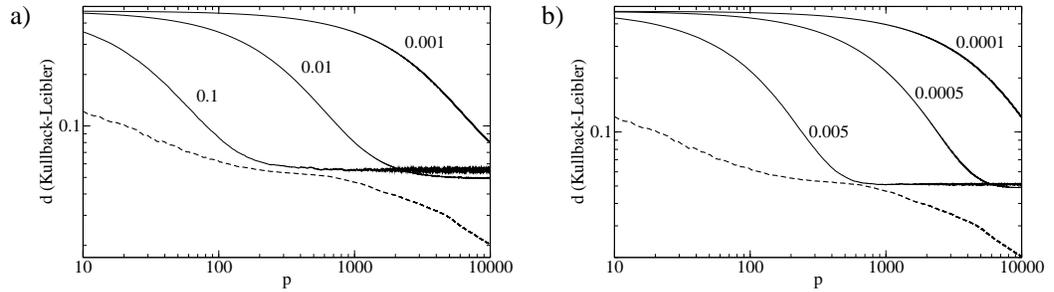}
\caption{a) Comparison between MPA (dashed) and BC (continuous). 
         Values of $\lambda$ are indicated next to the curves. $\eta_{BC}=0.5$. 
         b) Comparison between MPA (dashed) and BWO (continuous). 
         Values of $\eta_{BW}$ are indicated next to the curves. Both scales are log-log.}
\label{fBWO05}
\end{figure}

Figure \ref{fBWO06} compares MPA and BOnA. The initial
difference
decreases in time and both come
closer relatively fast.
We used $n=2$, $m=3$ and 
$T=2$ and averaged over 150 random teachers with symmetric initial 
students. The computational time for BOnA was 340min, 
and for MPA, 5s in a 1GHz processor. Figure \ref{fBWO05}a compares MPA to BC and figure 
\ref{fBWO05}b to BWO. In both 
cases MPA has better generalisation. We used $n=2$, $m=3$, $T=2$, symmetric initial students and 
averaged over 500 random teachers. 

\section{Learning Drifting Concepts}

\begin{figure}
\centering
\includegraphics[width=13.6cm]{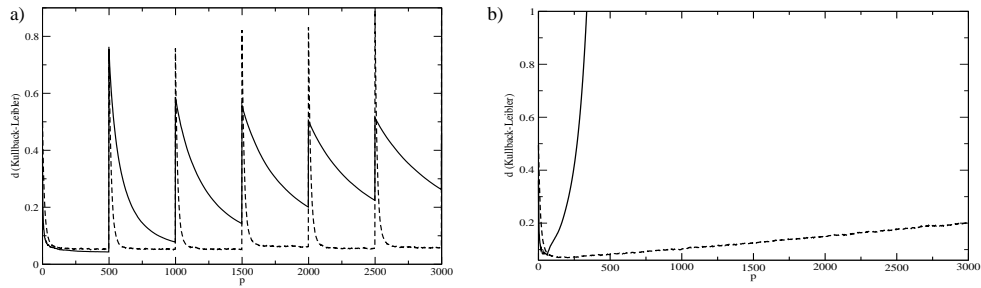}
\caption{Drifting concepts. Continuous lines correspond to MPA and dashed lines to BC. 
         a) Abrupt changes at 500 sequences interval. 
         b) Small random changes at each new sequence.}
\label{fLDC01}
\end{figure}

We tested BC and MPA for changing teachers. In figure \ref{fLDC01}a, it
changes at random after each 500 sequences ($\lambda=0.01$, $\eta_{BC}=10.0$). 
In figure \ref{fLDC01}b, each time a sequence is observed, a small 
random quantity is added to the teacher. Both have 
$n=2$, $m=3$ and are averaged over 200 runs.

Figure \ref{fLDC01}b shows that BC adapts better, but is not \emph{fully} adaptive and we do not know 
how to modify it. MPA instead derives from Bayesian principles and we can guess the problem by 
analogy with similar Bayesian algorithms \cite{Vicente}: variances 
decrease in the process as in the perceptron, where they are the 
learning rates, explaining the memory effect difficulting the learning after changes. 
Although not proved yet, 
we expect the same relationship in MPA, which can be used to improve performance.

\section{Learning and Symmetry Breaking}

Learning from symmetric initial students requires that the parameters separate 
from each other in some point, which
depends on the algorithm and is an important feature in \textit{online} 
algorithms \cite{Heskes}, breaking the symmetry with a sharp decrease in the 
generalisation error. 

Instead of taking averages to smooth abrupt 
changes, here we draw curves for only one teacher, rendering them visible. 
Flat lines before a symmetry breaking are called \textit{plateaux} and occur when it is 
difficult to break the symmetry.

\begin{figure}
\centering
\includegraphics[width=12.8cm]{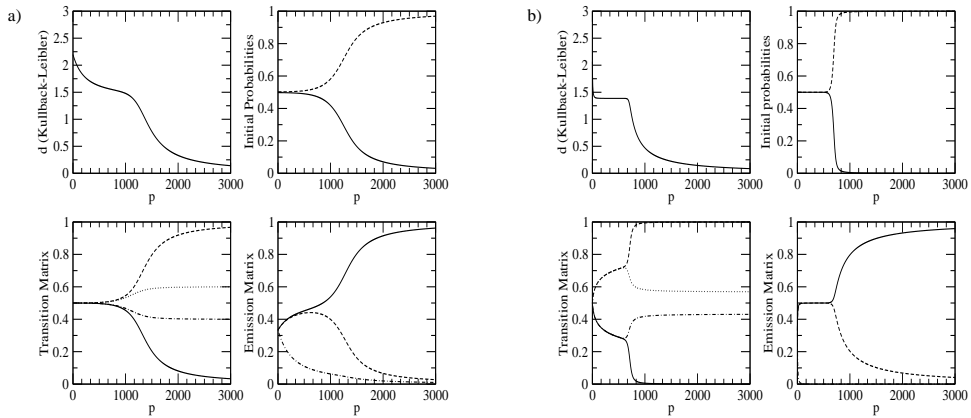}
\caption{KL-divergence and student's parameters for a) BC  and b) MPA.}
\label{gSB1}
\end{figure}

\begin{figure}
\centering
\includegraphics[width=5.6cm]{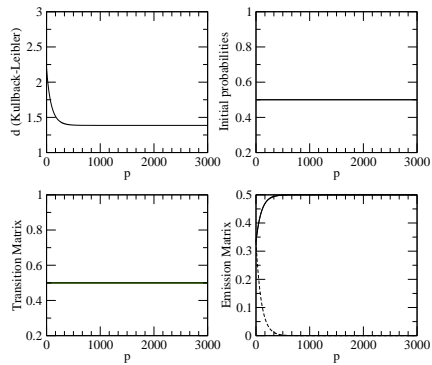}
\caption{KL-divergence and student's parameters for BWO.}
\label{gSB3}
\end{figure}

Figure \ref{gSB1}a shows BC ($\lambda=0.01$, $\eta_{BC}=1.0$) 
with two abrupt changes: in the
beginning and after 1000 sequences. $\pi$ and $A$ only break the symmetry
in the second point, and $B$ in both. Figure \ref{gSB1}b shows that
in MPA the second change is stronger and the symmetry breaking affects
both $B$ and $A$. Figure \ref{gSB3} shows BWO with
$\eta_{BW}=0.01$ where only $B$ is affected.
The more symmetries are broken, the best the generalisation of the algorithm.

In all simulations we set $n=2$, $m=3$ and $T=2$ with a teacher HMM given by
\begin{equation}
  \pi=
    \left(
    \begin{array}{c}
      1\\
      0
    \end{array}
    \right),
  \qquad
  A=	
    \left(
    \begin{array}{cc}
      0 & 1\\
      1 & 0
    \end{array}
    \right),
   \qquad
  B=	
    \left(
    \begin{array}{ccc}
      1 & 0 & 0\\
      0 & 0 & 1
    \end{array}
    \right).
\end{equation}

\section{Conclusions}
We proposed and analysed three learning algorithms for HMMs: Baum-Welch Online (BWO), 
Bayesian Online Algorithm (BOnA) and Mean Posterior Approximation (MPA). We showed the superior performance of MPA  
for static teachers, but the Baldi-Chauvin (BC) algorithm is better for drifting 
concepts, although the Bayesian nature of MPA 
suggests how to fix it. The results seem to be confirmed by initial tests on real data.

The importance of symmetry breaking in learning processes is presented here in
a brief discussion where the phenomenon is shown to occur in our models.

\section{Acknowledgements}
We would like to thank Evaldo Oliveira, Manfred Opper and Lehel Csato for 
useful discussions. This work was made part in the University of S\~ao Paulo with 
financial support of FAPESP and part in the Aston University with support of Evergrow Project.


\begin{thebibliography}{99}

\bibitem{Ephraim02} Y. Ephraim, N. Merhav, Hidden Markov Processes. IEEE Trans. Inf. Theory \textbf{48}, 1518-1569 (2002).

\bibitem{Rabiner89} L. R. Rabiner, A Tutorial on Hidden Markov Models and Selected Applications in Speech Recognition.
Proc. IEEE \textbf{77}, 257-286 (1989).

\bibitem{Baldi01} P. Baldi, S. Brunak, Bioinformatics: The Machine Learning Approach. MIT Press (2001).

\bibitem{Durbin98} R. Durbin, S. Eddy, A. Krogh, G. Mitchison, Biological sequence analysis: Probabilistic models of 
proteins and nucleic acids. Cambridge University Press, Cambridge (1998).

\bibitem{Frias04} J. Garcia-Frias, Decoding of Low-Density Parity-Check Codes Over Finite-State Binary Markov Channels.
IEEE Trans. Comm. \textbf{52}, 1840-1843 (2004).


\bibitem{Baldi94} P. Baldi, Y. Chauvin, Smooth On-Line Learning Algorithms for Hidden Markov Models. 
Neural Computation \textbf{6}, 307-318 (1994).

\bibitem{Amari} S. Amari, Neural learning in structured parameter 
spaces - Natural Riemannian gradient. NIPS'96 \textbf{9}, MIT Press (1996).

\bibitem{Opper98} M. Opper, A Bayesian Approach to On-line Learning. On-line learning in Neural Networks, 
edited by D. Saad, Publications of the Newton Institute, Cambridge Press, Cambridge (1998).

\bibitem{Saad} M. Opper, D. Saad, Advanced Mean Field Methods: Theory and Practice. MIT Press (2001).

\bibitem{Alamino} R. Alamino, N. Caticha, Bayesian Online Algorithms for Learning in Discrete Hidden Markov Models. 
Submitted to Discrete and Continuous Dynamical Systems.  

\bibitem{Heskes} T. Heskes, W. Wiegerinck, W., On-line Learning with
Time-Correlated Examples. On-line Learning in Neural Networks,
251-278, edited by David Saad, Cambridge University Press, Cambridge (1998). 

\bibitem{Vicente} R. Vicente, O. Kinouchi, N. Caticha. Statistical
Mechanics of Online Learning of Drifting Concepts: A Variational Approach.
Machine Learning \textbf{32}, 179-201 (1998).

\bibitem{Vlad} M. O. Vlad, M. Tsuchiya, P. Oefner, J. Ross. 
Bayesian analysis of systems with random chemical composition: 
Renormalization-group approach to Dirichlet distributions and the statistical
theory of dilution. Phys. Rev. E \textbf{65}, 011112(1)-01112(8) (2001).

\end{thebibliography}
\end{document}